%% file: main.tex
\documentclass{article}

\usepackage{microtype}
\usepackage{graphicx}
\usepackage{subfigure}
\usepackage{booktabs} %

\usepackage{hyperref}

\usepackage[accepted]{mlsys2024}
\input{macro}

\mlsystitlerunning{\input{content/000_title}}

\begin{document}

\twocolumn[
\mlsystitle{\input{content/000_title}}

\mlsyssetsymbol{equal}{*}

\begin{mlsysauthorlist}

\mlsysauthor{Xiaoyuan Liu}{ucb}
\mlsysauthor{Tianneng Shi}{ucb}
\mlsysauthor{Chulin Xie}{uiuc}
\mlsysauthor{Kangping Hu}{zju}
\mlsysauthor{Haoyu Kim}{pku}
\mlsysauthor{Xiaojun Xu}{uiuc}
\mlsysauthor{The-Anh Vu-Le}{uiuc}
\mlsysauthor{Zhen Huang}{sjtu}
\mlsysauthor{Arash Nourian}{amazon}
\mlsysauthor{Bo Li}{uiuc}
\mlsysauthor{Dawn Song}{ucb}
\end{mlsysauthorlist}

\mlsysaffiliation{ucb}{UC Berkeley}
\mlsysaffiliation{uiuc}{University of Illinois at Urbana-Champaign}
\mlsysaffiliation{zju}{Zhejiang University}
\mlsysaffiliation{pku}{Peking University}
\mlsysaffiliation{sjtu}{Shanghai Jiao Tong University}
\mlsysaffiliation{amazon}{Amazon Web Services}

\mlsyskeywords{Federated Learning, Federated Learning Frameworks, Benchmark}

\vskip 0.3in

\input{content/010_abstract}
]

\printAffiliationsAndNotice{}  %

\input{content/100_intro}

\input{content/200_related}

\input{content/320_system}

\input{content/510_exp}

\input{content/600_discuss}

\bibliography{ref}
\bibliographystyle{mlsys2024}

\clearpage
\appendix

\input{content/appendix/all}

\end{document}

%% file: macro.tex
\usepackage{xspace}
\usepackage{amsmath}
\usepackage{graphicx}
\usepackage{xcolor}
\usepackage{multirow}
\usepackage{makecell}
\usepackage{enumitem}
\usepackage{wrapfig}
\usepackage{pifont}  %
\usepackage{cleveref}
\newcommand{\circled}[1]{\ding{\numexpr171+#1\relax}}

\newif\ifsubmit
\submittrue

\newcommand{\reviewer}[3]{
  \expandafter\newcommand\csname #1\endcsname[1]{
    \ifsubmit
      \ignorespaces
    \else
      \textcolor{#3}{[#2: ##1]}
    \fi
  }
}

\reviewer{todo}{TODO}{red}
\reviewer{xiaoyuan}{Xiaoyuan}{brown}
\reviewer{tianneng}{Tianneng}{cyan}
\reviewer{bo}{Bo}{green}
\reviewer{lun}{Lun}{magenta}

\newcommand{\yes}{\ding{51}}%
\newcommand{\no}{\ding{55}}%

\newcommand{\sysname}{UniFed\xspace}

%% file: content/000_title.tex
\sysname: All-In-One Federated Learning Platform to Unify Open-Source Frameworks

%% file: content/010_abstract.tex
\begin{abstract}

Federated Learning (FL) has become a practical and widely adopted distributed learning paradigm. However, 
the lack of a comprehensive and standardized solution covering diverse use cases makes it challenging to use in practice. In addition, selecting an appropriate FL framework for a specific use case can be a daunting task.
In this work, we present \sysname, the first unified platform for standardizing existing open-source FL frameworks. 
The platform streamlines the end-to-end workflow for distributed experimentation and deployment,
encompassing 11 popular open-source FL frameworks. In particular, to address the substantial variations in workflows and data formats, \sysname introduces a configuration-based schema-enforced task specification, offering 20 editable fields. \sysname also provides functionalities such as distributed execution management, logging, and data analysis.

With \sysname, we evaluate and compare 11 popular FL frameworks from the perspectives of functionality, privacy protection, and performance, through conducting developer surveys and code-level investigation. We collect 15 diverse FL scenario setups (e.g., horizontal and vertical settings) for FL framework evaluation. This comprehensive evaluation allows us to analyze both model and system performance, providing detailed comparisons and offering recommendations for framework selection.
\sysname simplifies the process of selecting and utilizing the appropriate FL framework for specific use cases, while enabling standardized distributed experimentation and deployment. Our results and analysis based on experiments with up to 178 distributed nodes provide valuable system design and deployment insights, aiming to empower practitioners in their pursuit of effective FL solutions.
\end{abstract}

%% file: content/100_intro.tex
\section{Introduction}
\label{sec:intro}

Federated Learning (FL)~~\cite{mcmahan2017communication,kairouz2021advances} has become a practical and popular paradigm for training machine learning (ML) models, and
there are many existing open-source FL frameworks, {such as {TFF \cite{tff}}, Flower \cite{beutel2020flower}, FATE \cite{fate2021} and FedScale \cite{lai2022fedscale}}. 
However, unlike PyTorch \cite{paszke2019pytorch} and TensorFlow \cite{abadi2016tensorflow} for ML, currently, there is not a unified systematic solution that is maturely developed for most use cases. 
FL practitioners often face the challenge of selecting the most suitable solution for specific use cases, which requires significant exploration effort. In addition, it is usually challenging to interpret and analyze the output of different FL frameworks given their different output formats and functionalities.

To tackle these challenges, we build \sysname(\url{https://bit.ly/unifed}), a standardization platform to facilitate easy experimentation with open-source FL solutions and enable real-world deployments. 
The platform unifies different frameworks and streamlines a consistent end-to-end workflow covering most FL use cases.

\begin{figure*}[hbt]
\centering
    \includegraphics[width=\textwidth]{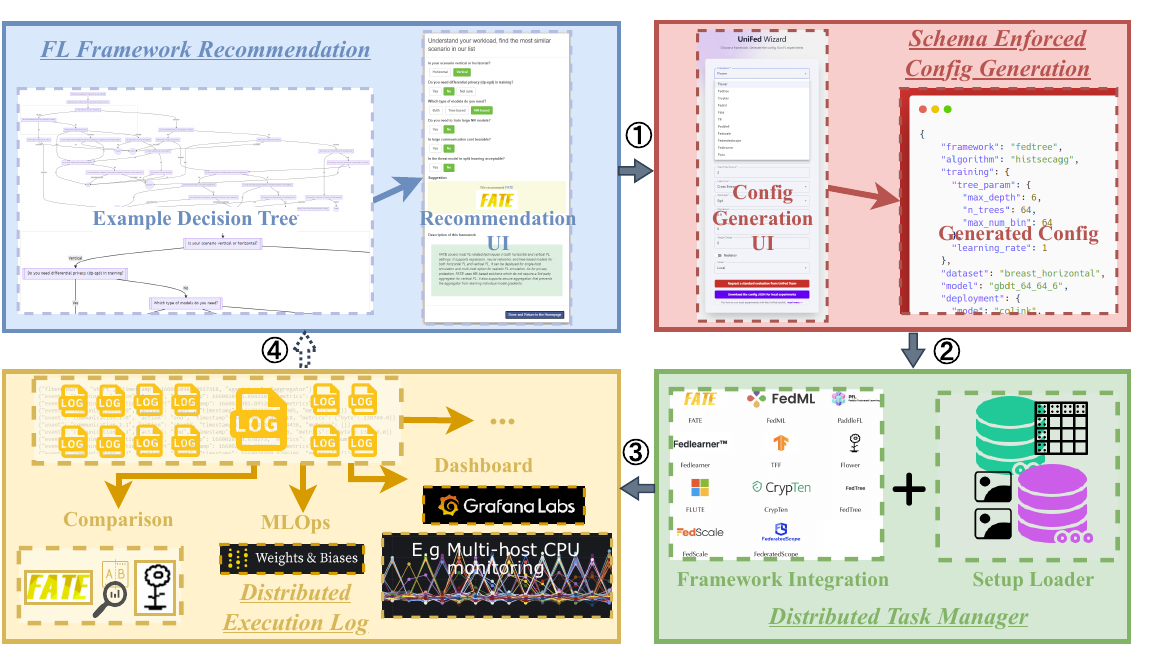}
      \caption{\small Overview of the \sysname platform. 
      \circled{1} For a given use case, \sysname provides recommendations for framework selection to guide the configuration generation with UI.
      \circled{2} With a consistent \sysname configuration, users can run distributed FL training with different frameworks and in different setups.
      \circled{3} During training, a distributed log is generated. Integration in \sysname with data analysis platforms enables convenient log investigations.
      \circled{4} Additionally, the analysis further improves the quality of framework recommendations.
      }
    \label{fig:intuition}
\end{figure*} 

Figure \ref{fig:intuition} provides an overview of the \sysname platform. 
A user first interacts with the framework recommendation UI to find potential matching frameworks. 
To run experiments, the user then fills in the form in the configuration generation UI. The generation UI provides detailed options about algorithm, model, and data source selection as well as training parameters. The UI also verifies the validity of the generated {config} file.
Providing the generated configuration file as the input,  \sysname  initiates a distributed training task using the specified framework and data source. 
In addition to getting the trained {FL} models, \sysname also generates distributed execution log in a consistent format as the output, which can be directly used for framework comparison. One can further investigate the log by connecting with other data analysis platforms integrated with \sysname.
Such an analysis reflects both the model and system performance of the selected framework, which can be further used to optimize the recommendation module, creating a closed loop for optimizing the user experience.

The contribution of this paper is summarized as follows.
(1) To understand the challenge of standardizing FL frameworks, we conduct developer surveys and code-level investigations to compare eleven existing open-source solutions, summarizing their functionalities, such as the supported FL algorithms and privacy protection mechanisms. 
(2) To address the substantial variations in workflows and data format, based on our observation of existing solutions, we built \sysname, a unified platform to standardize the input, execution, and output of different FL frameworks, streamlining the workflow for distributed FL experimentation and deployment consistently. 
(3) Based on \sysname, we evaluate the model and system performance of all eleven FL frameworks on a diverse collection of FL setups covering horizontal and vertical settings. We consolidate the comparison results and offer valuable insights to facilitate potential FL framework recommendations for different use cases.

%% file: content/200_related.tex
\section{Related work}
\label{sec:related_work}

\subsection{Existing FL datasets}
There are two kinds of federated datasets used in the experiments of FL studies: 1) simulated federated datasets by artificially partitioning a centralized dataset into multiple clients; 2) real federated datasets where the client datasets are collected from different sources. 
For the first category, existing FL studies \cite{mcmahan2017communication,luo2021no,fallah2020personalized,cheng2021secureboost,wu2020privacy} usually use the centralized image datasets (e.g., MNIST, CIFAR-10) or tabular datasets (e.g., credit) for partitioning due to their popularity and ease of use. 
FATE \cite{fate2021} provides a collection of simulated federated datasets varying from images, tabular data, to sensor data. For the second category, LEAF \cite{caldas2018leaf} collects six %
datasets that naturally fit the cross-device horizontal FL setting. For example, Sentiment140 in LEAF is a federated text dataset of tweets by considering each Twitter user as a client. However, as mentioned in \cite{kairouz2021advances}, there lack of open-source real feature-partitioned datasets for vertical FL.

We consider both simulated and real FL setups in diverse application domains. Specifically, considering the real-world usage of FL frameworks, we adopt the datasets  {from LEAF}~\cite{caldas2018leaf} to evaluate cross-device horizontal FL.
For cross-silo horizontal FL and vertical FL, we adopt datasets from FATE~\cite{fate2021}, covering representative applications from finance to healthcare.  

\subsection{Existing FL frameworks}

There are many system construction efforts on building \textit{FL frameworks} to support various FL scenarios.
In this work, we focus on open-source FL frameworks that are available for evaluation. Here we
identify three general categories along with representative examples and provide detailed comparisons in Section \ref{sec:config}.

\textit{All-in-one frameworks} cover most FL-related techniques in both horizontal and vertical FL settings. Such FL frameworks, including FATE \cite{fate2021}, FedML \cite{he2020fedml}, PaddleFL \cite{paddlefl}, Fedleaner \cite{fedlearner}, FederatedScope \cite{xie2023federatedscope}, and {TFF \cite{tff}}, focus on  {supporting different use-cases}. %

\textit{Horizontal-only frameworks} 
aim to provide easy-to-use APIs for users to develop horizontal FL algorithms, e.g., Flower \cite{beutel2020flower}, FLUTE \cite{flute}, FedScale \cite{lai2022fedscale}.

\textit{Specialized frameworks} are designed for specific purposes instead of general FL, including CrypTen \cite{crypten2020} and FedTree \cite{fedtree}.
CrypTen  provides secure multi-party computation \cite{yao1986generate} primitives, while FedTree is designed for FL with decision trees.

\subsection{Existing FL benchmarks}
Existing FL \textit{benchmarks} (1) mainly curate federated datasets from various domains and (2) develop own FL implementations (which may not be comprehensive enough for industrial purposes) to evaluate FL algorithms on their proposed datasets~\cite{caldas2018leaf,hu2020oarf, he2021fedgraphnn,lai2021fedscale,lin2021fednlp,li2022federated,liang2020flbench,chai2020fedeval}.
Specifically,
LEAF~\cite{caldas2018leaf} is a popular benchmark including 6 FL tasks on synthetic, CV and NLP data, but it only evaluates FedAvg. 
FLBench~\cite{liang2020flbench} proposes 9 federated datasets in medicine, finance, and IoT domains, yet without any empirical evaluation.
OARF~\cite{hu2020oarf} collects a total of 22 datasets on CV, NLP, and GIS tasks for horizontal FL and tabular datasets for vertical FL, and evaluates algorithm utility under differential privacy.
To generate non-IID FL datasets from existing centralized datasets, NIID-Bench~\cite{li2022federated} proposes 6 partitioning strategies and compares several FL algorithms on its partitioned datasets. 
FedEval~\cite{chai2020fedeval} compares FedAvg and FedSGD on existing FL datasets and provides privacy and robustness evaluation.
To accelerate the FL research on more domains and tasks, FedNLP~\cite{lin2021fednlp} creates 4 federated NLP datasets and FedGraphNN~\cite{he2021fedgraphnn} provides 36 federated graph datasets.
 FedScale~\cite{lai2021fedscale} incorporates 20 federated datasets spanning various tasks on CV, NLP, RL, and also a client system behavior dataset. FLINT~\cite{wang2023flint} discusses simulations of cross-device settings.
 
In summary, existing benchmarks mainly focus on creating federated datasets in different tasks, either from natural client data or from artificially partitioned centralized datasets, to evaluate FL \textit{algorithms}. 
However, they do not provide a unified platform to systematically evaluate different FL \textit{frameworks} that are built with industry efforts in practice. To fill in this gap, we build a unified platform \sysname, which provides the option to incorporate and compare eleven open-source FL frameworks with 15 common FL setups to cover different FL settings, data modalities, tasks, as well as workload sizes.

%% file: content/320_system.tex
\section{System design}
\label{sec:sys}

\begin{figure*}[hbt]
\centering
    \includegraphics[width=\linewidth]{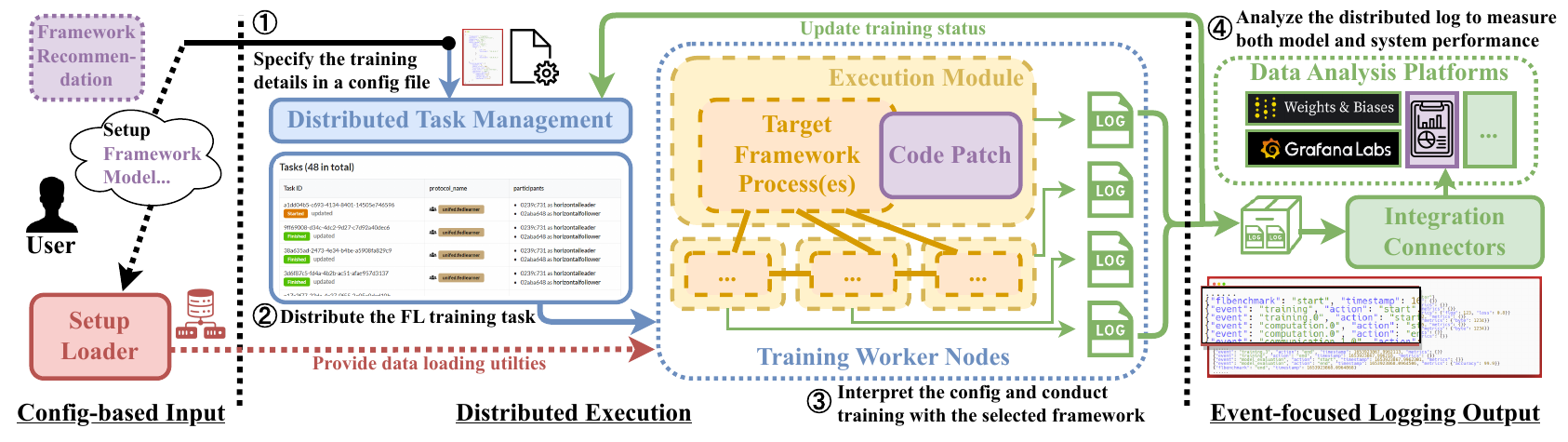}
      \caption{\small An end-to-end FL workflow across input, execution, and output stages facilitated by \sysname.
      }
    \label{fig:workflow}
\end{figure*} 

\subsection{Workflow overview} 
Integrating diverse FL frameworks poses a significant challenge, owing to the distinct usage and learning processes associated with each framework. To tackle this issue, \sysname implements a modular design that seamlessly facilitates the entire workflow of FL.
As illustrated in Figure \ref{fig:workflow}, the user starts by selecting a setup and a framework according to our recommendation, specifying the training details following the configuration format without prior knowledge about certain FL frameworks.
With the specification, the user connects with the training nodes through our distributed task management module and sends the well-formatted configuration file. The task management module distributes the task to the execution module on each node. The execution module interprets the configuration into specific operations using the selected setup and framework to conduct the training in a distributed way. After the task execution, each node generates a local log reflecting the training procedure also in a unified format. The log files are collected for further investigations with rich integration with different data analysis platforms.    

Through our modular design, \sysname unifies different FL frameworks across \textit{input}, \textit{execution}, and \textit{output} stages. 
To unify the input stage, \sysname provides a setup loading module with rich example setups and a consistent configuration format. 
To unify the execution stage, \sysname incorporates multi-host task management and node execution modules to simplify distributed evaluation.
To unify the output stage, \sysname designs an event-focused logging format compatible with existing data visualization and MLOps platforms~\cite{bonawitz2017practical}.
Analyzing the comparison experiment logs, \sysname also provides framework recommendations based on given requirements from real-world applications.

\input{content/330_workload}  %
\input{content/340_frameworks}  %
\input{content/350_management}  %
\input{content/360_output}

%% file: content/330_workload.tex
\subsection{Input stage: example setup collection and the setup loading module}
\label{sec:workload}

FL setups vary in participant setting, input data modality, and task types.
\sysname supports setups in both horizontal and vertical settings, depending on whether the data split among participants is on instances or features. Specifically, for horizontal settings, it supports both cross-silo and cross-device setups, with scales from two participants to hundreds.
As discussed in Section \ref{sec:related_work}, we collect 15 representative setups for evaluation. Our collection covers various ML task types and input data modalities of tabular, image, text, and others. We summarize the characteristics in Table~\ref{tab:scene}.

\input{content/fig/scenarios}

\sysname provides a setup loading module for accessing the setup collection, facilitating experiments with different frameworks and training parameter selections.
It automates the source dataset download and caching, splits datasets with a deterministic partition across different clients to stably generate setups for evaluation reproducibility, and is compatibly usable in different FL framework implementations for easy integration. 
The setup loading module is also easily customizable to add more example setups or connect with real-world data sources.

%% file: content/fig/scenarios.tex
\begin{table*}[h]
  \centering
  \resizebox{\textwidth}{!}{%
    \begin{tabular}{lllllcc}
      \toprule
      \textbf{\makecell{Setting}}
      & \textbf{\makecell{Setup name}} & \textbf{\makecell{Modality}} & \textbf{\makecell{Task type}} & \textbf{\makecell{Performance\\metrics}} & \textbf{\makecell{{Number of}\\{clients}}} & \textbf{\makecell{{Number of}\\{samples}}} \\
      \hline
      \multicolumn{1}{c|}{\multirow{5}{*}{\makecell{cross-device\\horizontal}}} & celeba \cite{liu2015deep} & Image & \makecell[l]{Binary Classification\\(Smiling vs. Not smiling)} & Accuracy & 894 & 20,028 \\
      \cline{2-7}
      \multicolumn{1}{c|}{} & femnist \cite{mnist,cohen2017emnist,caldas2018leaf} & Image & \makecell[l]{Multiclass Classification\\(62 classes)} & Accuracy & 178 & 40,203 \\
      \cline{2-7}
      \multicolumn{1}{c|}{\makecell{\\\\}} & reddit \cite{caldas2018leaf}& Text & Next-word Prediction & Accuracy & 813 & 27,738 \\
      \hline
      \multicolumn{1}{c|}{\multirow{9}{*}{\makecell{cross-silo\\horizontal}}} & breast\_horizontal \cite{breast} \makecell{\\\\} & Medical & Binary Classification & AUC & 2 & 569 \\
      \cline{2-7}
      \multicolumn{1}{c|}{\makecell{\\\\}} & default\_credit\_horizontal \cite{yeh2009comparisons,Dua:2019} & Tabular & Binary Classification & AUC & 2 & 22,000 \\
      \cline{2-7}
      \multicolumn{1}{c|}{\makecell{\\\\}} & give\_credit\_horizontal \cite{givecredit} & Tabular & Binary Classification & AUC & 2 & 150,000 \\
      \cline{2-7}
      \multicolumn{1}{c|}{} & student\_horizontal \cite{cortez2008using,Dua:2019} & Tabular & \makecell[l]{Regression\\(Grade Estimation)} & MSE & 2 & 395 \\
      \cline{2-7}
      \multicolumn{1}{c|}{} & vehicle\_scale\_horizontal \cite{siebert1987vehicle,Dua:2019} & Image & \makecell[l]{Multiclass Classification\\(4 classes)} & Accuracy & 2 & 846 \\
      \hline
      \multicolumn{1}{c|}{} & & & & & \multicolumn{2}{c}{\textbf{Vertical split details}} \\
      \multicolumn{1}{c|}{\multirow{13}{*}{vertical}} & breast\_vertical \cite{breast} \makecell{\\\\} & Medical & Binary Classification & AUC & \multicolumn{2}{l}{\makecell[l]{A: 10 features 1 label\\B: 20 features}}\\
      \cline{2-7}
      \multicolumn{1}{c|}{\makecell{\\\\}} & default\_credit\_vertical \cite{yeh2009comparisons,Dua:2019} & Tabular & \makecell[l]{Binary Classification} & AUC & \multicolumn{2}{l}{\makecell[l]{A: 13 features 1 label\\B: 10 features}}\\
      \cline{2-7}
      \multicolumn{1}{c|}{\makecell{\\\\}} & dvisits\_vertical \cite{cameron1988microeconometric} & Tabular & \makecell[l]{Regression (Number of\\consultations Estimation)} & MSE & \multicolumn{2}{l}{\makecell[l]{A: 3 features 1 label\\B: 9 features}}\\
      \cline{2-7}
      \multicolumn{1}{c|}{\makecell{\\\\}} & give\_credit\_vertical \cite{givecredit} & Tabular & Binary Classification & AUC & \multicolumn{2}{l}{\makecell[l]{A: 5 features 1 label\\B: 5 features}}\\
      \cline{2-7}
      \multicolumn{1}{c|}{\makecell{\\\\}} & motor\_vertical \cite{motor} & Sensor data & \makecell[l]{Regression\\(Temperature Estimation)} & MSE & \multicolumn{2}{l}{\makecell[l]{A: 4 features 1 label\\B: 7 features}}\\
      \cline{2-7}
      \multicolumn{1}{c|}{\makecell{\\\\}} & student\_vertical \cite{cortez2008using,Dua:2019} & Tabular & \makecell[l]{Regression\\(Grade Estimation)} & MSE & \multicolumn{2}{l}{\makecell[l]{A: 6 features 1 label\\B: 7 features}}\\
      \cline{2-7}
      \multicolumn{1}{c|}{\makecell{\\\\}} & vehicle\_scale\_vertical \cite{siebert1987vehicle,Dua:2019} & Image & \makecell[l]{Multiclass Classification\\(4 classes)} & Accuracy & \multicolumn{2}{l}{\makecell[l]{A: 9 features 1 label\\B: 9 features}}\\
      \hline
    \end{tabular}%
  }
  \caption{\small Evaluation setups in \sysname. \sysname borrow $15$
    datasets from existing works to cover different FL settings, modalities, task types, and workload sizes.
  }
  \label{tab:scene}
\end{table*}

%% file: content/340_frameworks.tex
\subsection{Input stage: unified configuration interface with enforcement}
\label{sec:config}

\sysname allows users to access rich functionalities in 11 different open-source FL frameworks.
These frameworks have different system designs, use different libraries, and target different privacy protection goals. As a result, there are huge differences in how they operate. We first investigate the functionality support in these FL frameworks. Building upon our observations of the accessible choices and parameters, we proceed to design a universal configuration interface that enables the definition of an FL task capable of running on any of these frameworks.

To investigate the functionality support, we conduct surveys targeting framework authors and maintainers and also explore the implementation ourselves to get first-hand experience.
We designed a questionnaire form with 18 
questions reflecting the framework support from the perspectives of model types, deployment, privacy protection, and utility. 
We sent out the questionnaires to the maintainer teams of all 11 frameworks and received 5 official replies covering around a hundred critical answers.
The design of the questionnaire form and detailed discussions are in Appendix \ref{sec:docreq}. 
To confirm the survey results and get further details, we explore the open-source implementations and evaluate their support for different models in different participant settings. %
Based on the survey and verified by our integration and testing, we summarize our findings about the functionality support of different frameworks in \Cref{tab:framework} and the usability features of different frameworks in \Cref{tab:usability}.

\input{content/fig/function}

\input{content/fig/usability}

Based on our observations of available functionalities, we design a configuration schema to reflect options provided by different FL frameworks.
To run FL tasks, \sysname users only need to create a configuration JSON file, and \sysname system automates the rest of the procedure following the configuration as the blueprint. 
Furthermore, \sysname also provides wizard UI\footnote{\url{https://bit.ly/unifed-wizard}} to guide the users in configuration file generation, which is as simple as filling an interactive form as shown in Figure \ref{fig:intuition}. 
Specifically, the JSON-formatted configuration includes 6 main fields and 14 optional fields, covering selections of the framework, training algorithm, model architecture, evaluation setup, deployment mode, and various training parameters. 
Leveraging a structural schema definition file, the automatically generated UI not only reduces the human learning effort but also provides strong enforcement to prevent malformed configurations with invalid value combinations. 

%% file: content/fig/function.tex
\begin{table*}[h]
    \centering
\resizebox{\textwidth}{!}{%
\begin{tabular}{lccccccccccc}
\toprule
\multicolumn{1}{c|}{\multirow{2}{*}{\textbf{Framework}}} & \multicolumn{6}{c|}{\textbf{All-in-one frameworks}} & \multicolumn{3}{c|}{\textbf{Horizontal-only frameworks}} & \multicolumn{2}{c}{\textbf{Specialized frameworks}}\\
\multicolumn{1}{c|}{} & \textbf{FATE} & \textbf{FedML} & \textbf{PaddleFL} & \textbf{Fedlearner} & \textbf{{FederatedScope}} & \multicolumn{1}{c|}{\textbf{TFF}} & \textbf{Flower} & \textbf{FLUTE} & \multicolumn{1}{c|}{\textbf{{FedScale}}} & \textbf{CrypTen} & \textbf{FedTree} \\
\midrule
\multicolumn{11}{l}{\textbf{Model support - Horizontal}} \\
Regression & \yes & \yes & \yes & \yes & \yes & \yes & \yes & \yes & \yes & N/A & \no \\
Neural network & \yes & \yes & \yes & \yes & \yes & \yes & \yes & \yes & \yes & N/A & \no \\
Tree-based model & \yes & {\yes*} & \no & \no & \no & \no & \no & \no & \no & N/A & \yes \\
\midrule
\multicolumn{11}{l}{\textbf{Model support - Vertical}} \\
Regression & \yes & \yes & \yes* & \no & \yes & {\yes*} & \no & \no & \no & \yes & \no \\
Neural network & \yes & {\yes} & \yes* & \yes & \no & {\yes*} & \no & \no & \no & \yes & \no \\
Tree-based model & \yes & {\yes*} & \no & \yes & {\yes} & \no & \no & \no & \no & \no & \yes \\
\midrule
\multicolumn{11}{l}{\textbf{Deployment support}} \\
Single-host simulation &  \yes & \yes & \yes & \yes & \yes & \yes & \yes & \yes & \yes & \yes & \yes \\
Multi-host simulation on one machine & \yes & \yes & \yes & \yes & \yes & \no & \yes & \yes & \yes & \yes & \yes \\
Multi-host simulation on multiple machines & \no & \yes & \yes & \yes & \yes & \no & \yes & \yes & \yes & N/A & \yes \\
Edge-devices deployment (e.g., mobile) &  \no & \yes & \no & \no & \no & \yes & \no & \no & \no & \no & \no \\
{Networking protocols} & {Customized} &\makecell{MPI, gRPC, \\Torch Distributed, etc.} & {gRPC} & {gRPC} & {gRPC} & {gRPC} & {gRPC} & \makecell{MPI, \\ Torch Distributed} & {gRPC} & {Torch Distributed} & {gRPC} \\
\midrule
\multicolumn{11}{l}{\textbf{Privacy protection against the semi-honest server}} \\
Does not require a third party aggregator & \yes & \yes & \yes & \yes & \no & \yes & \no & \no & \no & \yes & \yes \\
Aggregator does not access to individual gradient/update & \yes & \yes & \yes & \no & \yes & \yes & \no & \no & \yes & N/A & {\no} \\
\midrule
\multicolumn{11}{l}{\textbf{Privacy protection against semi-honest peer clients}} \\
Clients do not learn gradients from other clients  & \yes & \yes & \yes & N/A & \yes & N/A & N/A & N/A & {N/A} &  \yes & \yes \\
Model params are partially revealed to clients & \yes & \yes & \yes & \no & \yes & \yes & N/A & \no & {N/A} & \yes & \yes \\
\midrule
\multicolumn{11}{l}{\textbf{Privacy protection in the FL model}} \\
Private training mechanisms (e.g. DPSGD) & \no & \yes & \yes & \no & \yes & \yes & \yes & \yes & \yes & \no & \yes \\
\midrule
\multicolumn{11}{l}{\textbf{Utility}} \\
GPU support & \yes & \yes & \yes & \yes & \yes &  \yes & \yes & \yes & \yes &  \yes & \no \\
Rich Optimizers & Customized & \yes &  Customized & \yes &  \yes &  \yes &  \yes &  \yes &  \yes &  Only SGD & \no \\
{ML backend}& {PyTorch, TF} & {PyTorch} & {PaddlePaddle} & {TF} & {PyTorch, TF} & {TF, JAX} & {PyTorch, TF, JAX, etc. } & {PyTorch} & {PyTorch, TF} & {PyTorch} & {
scikit-learn} \\
\bottomrule
\end{tabular}%
}
\caption{\small Functionality support in different FL frameworks. Asterisks{(*)} indicate a claimed support for certain functionalities that are missing or cannot run in the open-source implementation.  TF stands for ``TensowFlow''.
}
\label{tab:framework}
\end{table*}

%% file: content/fig/usability.tex
\begin{table*}
    \centering
\resizebox{\textwidth}{!}{%
\begin{tabular}{lccccccccccc}
\toprule
\multicolumn{1}{c|}{\multirow{2}{*}{\textbf{Framework}}} & \multicolumn{6}{c|}{\textbf{All-in-one frameworks}} & \multicolumn{3}{c|}{\textbf{Horizontal-only frameworks}} & \multicolumn{2}{c}{\textbf{Specialized frameworks}}\\
\multicolumn{1}{c|}{} & \textbf{FATE} & \textbf{FedML} & \textbf{PaddleFL} & \textbf{Fedlearner} & \textbf{{FederatedScope}} & \multicolumn{1}{c|}{\textbf{TFF}} & \textbf{Flower} & \textbf{FLUTE} & \multicolumn{1}{c|}{\textbf{{FedScale}}} & \textbf{CrypTen} & \textbf{FedTree} \\
\midrule
\multicolumn{11}{l}{\textbf{Documentation}} \\
Detailed tutorial & \yes & \yes & \yes & \no &  \yes &  \yes & \yes & \yes & \yes &  \yes & \yes \\
Code example & \yes & \yes & \yes & \yes & \yes &  \yes & \yes & \yes & \yes &  \yes & \yes \\
API documentation &  \yes & \no &  \no &  \no &  \yes &  \yes & \yes & \no &  \yes &  \yes & \yes \\
Native test \& benchmark & \yes & \yes & \no &  \no &  \yes &  \yes & \yes & \yes & \yes &  \yes & \yes \\
\midrule
\multicolumn{11}{l}{\textbf{Built-in ML building block}} \\
CNN &  \yes & \yes & \yes & \yes & \yes &  \yes & \yes & \yes & \yes &  \yes & \no \\
Transformer &  \yes & \yes & \yes & \yes & \yes &  \yes & \yes & \yes & \yes &  \no &  \no \\
\bottomrule
\end{tabular}%
}
\caption{\small Usability feature comparison in different FL frameworks. 
}
\label{tab:usability}
\end{table*}

%% file: content/350_management.tex
\subsection{Execution stage: distributed task management and node execution modules}
\label{sec:task}

FL involves asymmetric computation procedures on multiple machines. Managing such distributed computation introduces significant challenges. Existing FL frameworks employ different workflows for distributed training. Some of them focus on simulation only and their workflows require further wrapping for real-world deployment. \sysname bridges the gap between different user experiences among frameworks and \textit{provides necessary system features to support realistic deployment}.

Specifically, in \sysname, we identify three major challenges for managing distributed FL training in a realistic setting: \textit{multiple execution separation}, \textit{dynamic committee formation}, and \textit{bridging framework differences}. 
First, many frameworks handle one training run by default, providing scripts for a single execution without organizing the output result. However, real-world FL training often involves the reuse of the local data across multiple runs and requires separation in their computation procedure.
Second, many frameworks assume known participants on known machines before execution, while in the real world, the committee is often decided by the instantiation of the training task rather than before the system set-up.
Last, although \sysname configuration provides a consistent way to define an FL task, there still exist great differences between the setup procedure and operation in different FL frameworks. Realizing the specified training procedure consistently across different FL frameworks still requires significant effort.

\sysname incorporates a distributed task management module and a node execution module to tackle the above challenges.
The distributed task management module provides APIs on a command node to connect with participant nodes, while the execution management module is deployed on each participant node.
Specifically, we employ three key designs in response to the challenges in realistic FL training.
To support multiple training procedures and provide separation, in the distributed task management module, we pack an FL training procedure as an asynchronous distributed task and employ task management: assigning task IDs, managing their states, and queuing the computation request. Each FL training task is defined by a configuration file discussed in Section \ref{sec:config}. Then later in the node execution module, training runs with different task IDs are conducted in the separated space.
To support customized set-up for specific frameworks and form the committee dynamically, in the node execution module, we add a pre-training step for participants to exchange communication address information and prepare the framework execution environment.
Then during the execution, the node execution module interprets the configuration and translates it into operations following the workflow in the selected FL framework. The distributed deployment of the two modules free \sysname users from digging into details of how specific framework operates, allowing them to focus on the important model training and parameter tuning.

%% file: content/360_output.tex
\subsection{Output stage: event-focused logging with analysis platform integration}

In addition to unifying the input of the configuration file in Section \ref{sec:config}, and the training process in Section \ref{sec:task}, \sysname also aligns the output of the FL training and simplifies its analysis.
The output of an FL training procedure usually consists of two parts: the model checkpoints and the meta-information about the training procedure. In \sysname, we keep the model checkpoints generated by FL frameworks as is and focus on the meta-information. This is because the trained models are usually already in good shape for being deployed with the original framework. In addition, there are existing orthogonal efforts (e.g. \cite{bai2019}) bridging the gap between different model formats.

In practice, the meta-information about the FL training procedure is often reported in both the standard output of the training process and the logging files. However, even though different FL frameworks offer the same type of information, the information is often provided in different workflows and formats. For direct and consistent comparison, \sysname designs its own format for tracking meta-information output. Specifically, it uses event-focused distributed logging to track the training details reflecting both the model performance (AUC, MSE, and accuracy) and system performance (training time, communication size, and memory usage).

The design to handle output from different FL frameworks should satisfy three requirements: distributed, expressive, and comprehensive. As the training procedure is on multiple machines, distributed tracking with views from all participants is informative for performance analysis. Considering various FL algorithms involve different protocol steps and various metrics should be tracked, the output format needs to be expressive and flexible. Not all frameworks provide all the required information for a fair comparison, to conduct a comprehensive analysis, an additional mechanism is required for adding the missing and unaligned information.

To match the requirements, \sysname augments its node execution module with both the training and a post-training stage. First, to provide separate views from different participants, each node generates its local log and later sends it back to the command node to form a group of distributed logs. Second, to reflect the changes in training and protocol steps and report both computation and communication-related consumption, the log records the timestamped training event start and end with relevant metrics. Finally, to generate such logs, if the off-the-shelf FL framework has already provided the required information, we aligned the information to our format by adding a conversion in the post-training stage. However, if the framework itself does not track the required information, following a minimal modification principle, we create patches targeting specific frameworks to generate the required log. We summarize the main modifications of our patch in Appendix \ref{sec:patch}.

With our structural and expressive logging format, \sysname also facilitates convenient integration with data analysis platforms.
For example, we build connector modules to provide native support to existing MLOps and visualization platforms, porting our format to match their APIs. The MLOps platform helps provide easy model performance comparisons tracking the training parameter selection, while the visualization platform provides detailed illustrations for profiling the distributed procedure and analyzing system performance.
With \sysname, FL practitioners can easily compare different FL frameworks and gain insights into both model and system performance on various training setups using different frameworks.

%% file: content/510_exp.tex
\section{Experiment}
\label{sec:exp}

\sysname facilitates easy FL experimentation and deployment with a rich framework, algorithm, and model support. 
While our main goal is to provide a platform for FL researchers and practitioners to use, in this section, we conduct experiments based on \sysname to show its ability to run and compare different FL frameworks. \sysname supports metrics for different tasks, including AUC for binary classification, MSE for regression, and accuracy for multi-class classification. Also, \sysname includes important factors to reflect the system performance including training efficiency, communication cost, and resource consumption. Based on our experiments, we summarize insights to guide the selection and configuration of frameworks. For conciseness, here we present the key insights and representative experimental results in the main paper and put additional experiment results in Appendix \ref{sec:addexp}.

Not all selected frameworks support distributed deployment. Furthermore, for the ones that allow distributed evaluation, some of them only support simulators on worker nodes that act as different clients in different rounds.
To measure model performance, simulations are sufficient as they usually perform the same computation procedure as the distributed version.\footnote{There are also rare cases where the model performance in simulation is different from the one in deployment due to implementation details. We have provided corresponding feedback to the framework developers.}
However, the comparison of system performance is only meaningful when the frameworks are in deployment mode. 
With such considerations, in the rest of this section, we first discuss our implementation and evaluation environment. 
Then, we discuss our experiment in different settings to provide insights on framework usage and configuration selection. %
Moreover, through experiments, it becomes apparent that relying solely on a single framework often imposes limitations on functionality. The versatility offered by \sysname allows one to seamlessly experiment with multiple frameworks while adhering to a unified workflow, facilitating convenient exploration for the optimal choice for each encountered setup.

\input{content/520_impl}

\input{content/530_silo_h}

\input{content/540_silo_v}

\input{content/550_device_h}

%% file: content/520_impl.tex
\subsection{Implementation}

We implement and have released \sysname system discussed in Section \ref{sec:sys}. 
Specifically, the distributed task management module uses gRPC~\cite{grpc} to communicate task status and configurations with execution modules. We wrap the data loading for datasets from \cite{fate2021}, further automate the file caching from \cite{caldas2018leaf}, and fix the dataset splits for the evaluation setup following the procedures and parameters in the previous works.
For each framework, we create a separate code patch following the principle of minimal intrusion and resource consumption. We explain the details about separate patches for frameworks in Appendix \ref{sec:patch}.
We implement W\&B~\cite{wandb} and Grafana~\cite{grafana} connectors to support potential further analysis.
We use node exporter in Prometheus~\cite{prometheus} to collect certain system metrics. 
An example of CPU, memory, and network IO monitoring for distributed environment is provided in Appendix \ref{sec:grafana}.

\paragraph{Machines}
All cross-silo experiments use AWS r6i.2xlarge instances with 8 vCPU (2.90GHz) and 64 GiB RAM, and cross-device experiments with AWS r6i.xlarge instances with 4 vCPU (2.90GHz) and 32 GiB RAM.
All instances are running Ubuntu 20.04 LTS.

\paragraph{Hyperparameters}
For femnist dataset, we randomly select a subset of clients out of 178 clients at each round.  Each client uses the SGD algorithm with a batch size of 8, a learning rate of 0.001, and a momentum of 0.9 to update its local model for 1 local epoch. We run the FL algorithm for 2000 rounds to converge. 
For give\_credit\_horizontal dataset, we select all clients and update the local models with SGD for 1 local epoch. The batch size is 64, the learning rate is 0.01 and the momentum is 0.9, and we run 30 rounds. 
In addition, for default\_credit\_vertical dataset, we use a batch size of 128, a learning rate of 0.01, and a momentum of 0.9 to update its local model for 1 local epoch. We run the FL algorithm for 10 rounds to converge.
Similarly, for breast\_horizontal dataset, every client runs the SGD algorithm with a batch size of 32, a learning rate of 0.01, and a momentum of 0.9 to update its local model for 1 local epoch. We run 30 rounds for the FL algorithm to converge.
For default\_credit\_horizontal we use a batch size of 64, a learning rate of 0.01, and a momentum of 0.9 to update its local model for 1 local epoch. We run the FL algorithm for 30 rounds to converge.
We refer the readers to Appendix \ref{sec:exp-details} for versions of the evaluated FL frameworks.

%% file: content/530_silo_h.tex
\subsection{Cross-silo horizontal setting: the model selection dominates the framework selection}
\label{sec:csh}

\begin{table*}[hbtp]
    \centering
\resizebox{\textwidth}{!}{%
\begin{tabular}{c|cccccccccc}
\toprule
\textbf{\makecell{Model}} &   \textbf{\makecell{FedML}} & \textbf{\makecell{Flower}} & \textbf{\makecell{FLUTE}} & \textbf{\makecell{{FederatedScope}}} & \textbf{\makecell{{FedScale}}} & \textbf{\makecell{FATE}} & \textbf{\makecell{TFF}} & \textbf{\makecell{Fedlearner}} & \textbf{\makecell{PaddleFL}}\\
\hline
 \multicolumn{1}{l}{Regression} &  
 78.85$\pm$  0.46 &
 73.21$\pm$ 	5.44 & 
 76.11$\pm$	5.55 &
 77.93$\pm$1.11 &
 78.53$\pm$		0.00& 
 67.36$\pm$0.67 & 
 77.56$\pm$ 1.40 & 
 78.56 $\pm$	0.68 &
77.75 $\pm$ 0.08 
 \\
 \multicolumn{1}{l}{SLP} &  83.07	$\pm$ 0.10 &
82.93 $\pm$  	0.39 & 
83.02	$\pm$ 0.08   & 
83.12	$\pm$  0.08 & 
82.96	$\pm$ 0.00 & 
82.81$\pm$0.16 & 
83.56 	$\pm$ 0.66 & 
83.19 $\pm$	0.11 &
82.82 $\pm$	0.01 \\
 \multicolumn{1}{l}{3-layer MLP} & 
 {83.37$\pm$0.03}  & {83.51$\pm$0.06}  & {83.42$\pm$0.08}  & {83.39$\pm$0.08}  & {83.31$\pm$0.00}  & {83.01$\pm$0.08}  & {83.31$\pm$0.70}  & {83.39$\pm$0.16}  & {82.73$\pm$0.02}\\
\hline
\end{tabular}%
}
\caption{\small Model performance comparison across different frameworks supporting FedAvg in the setup ``give\_credit\_horizontal''. Most framework implementations achieve similar model performance.}
\label{tab:silohnn}
\end{table*}

Cross-silo horizontal setups often involve a few participants with different sets of data instances but the same feature set.
As the simplest among the three settings in terms of the system orchestration and communication pattern, the implementations of training algorithms from different frameworks are often similarly efficient.
In our experiment, we observed consistently good system performance for all candidates, where the convergence is usually within ten minutes. 
For model performance, we observe that the selection of model is decisive, similar to centralized ML training.

We first observe that when fixing the algorithm and model configuration, the framework selection does not introduce observable differences in the final model performance. Table \ref{tab:silohnn} show representative comparisons for the use of FedAvg on regression and NN-based modelsamong different frameworks. In Table \ref{tab:silohnn}, the difference between 3-layer MLP model performance is within 0.9\%. For Tree-based models, in our evaluation, Fate and FedTree achieve the same AUC (86.10). %

Isolating the factor of framework selection, we further observe that, similar to centralized training, to get better model performance, one should carefully choose the model type.
Table \ref{tab:rank} compares the NN-based model and tree-based model among the five setups in our collection. For results from different frameworks, we choose the best one available for comparison here. As we can observe, the tree-based model achieves better performance in most setups except for the "vehicle\_scale\_horizontal" with a small difference. Similar to centralized ML training, the tree-based model demonstrates an advantage as in our FL setups, the feature numbers are limited. %

\begin{table*}[htbp]
    \centering
\resizebox{\textwidth}{!}{%
\begin{tabular}{cccccccc}
\toprule
\multicolumn{1}{c|}{\multirow{2}{*}{\textbf{\makecell{Name}}}} & \multicolumn{2}{c|}{\textbf{1st}} & \multicolumn{2}{c|}{\textbf{2nd}} & \multicolumn{2}{c}{\textbf{3rd}} \\
\multicolumn{1}{c|}{} & \textbf{\makecell{alg\&model}} & \multicolumn{1}{c|}{\textbf{\makecell{perf}}} & \textbf{\makecell{alg\&model}} & \multicolumn{1}{c|}{\textbf{\makecell{perf}}} & \textbf{\makecell{alg\&model}} & \textbf{\makecell{perf}} \\
\hline
\makecell{breast\_horizontal\\(AUC)} & \makecell{HistSecAgg\\{gbdt\_64\_64\_6} } & {100.00$\pm$0.00\%} & \makecell{FedAvg\\{mlp\_128\_128\_128}} & {98.74$\pm$0.26\%} & \makecell{FedAvg\\mlp\_128} & {98.69$\pm$0.14\%}\\
\hline
\makecell{default\_credit\_horizontal\\(AUC)} & \makecell{HistSecAgg\\gbdt\_64\_64\_6} & {77.72$\pm$0.00\%}  & \makecell{FedAvg\\mlp\_128\_128\_128} & {77.62$\pm$0.14\%} & \makecell{FedAvg\\mlp\_128} & {76.49$\pm$0.56\%}\\
\hline
\makecell{give\_credit\_horizontal\\(AUC)} & \makecell{HistSecAgg\\gbdt\_64\_64\_6} & {86.10$\pm$0.00\%} & \makecell{FedAvg\\mlp\_128} & {83.57$\pm$0.06\%} & \makecell{FedAvg\\mlp\_128\_128\_128} & {83.51$\pm$0.60\%}\\
\hline
\makecell{student\_horizontal\\(MSE)} & \makecell{HistSecAgg\\{gbdt\_64\_64\_6}} & {22.56$\pm$0.98} & \makecell{FedAvg\\{mlp\_128\_128\_128}} & {23.16$\pm$0.44} & \makecell{FedAvg\\{mlp\_128}} & {23.41$\pm$0.39}\\
\hline
\makecell{vehicle\_scale\_horizontal\\(Accuracy)} & \makecell{FedAvg\\mlp\_128\_128\_128} & {100.00$\pm$0.00\%} & \makecell{FedAvg\\mlp\_128} & {100.00$\pm$0.00\%} & \makecell{HistSecAgg\\gbdt\_64\_64\_6} & {99.06$\pm$0.00\%}\\
\hline
\end{tabular}%
}
\caption{\small Model performance comparison in different setups. The top 3 model algorithm pairs are reported.}
\label{tab:rank}
\end{table*}

%% file: content/540_silo_v.tex
\subsection{Cross-silo vertical setting: the algorithm selection leads to significant system trade-offs}

The cross-silo vertical setting often involves two participants. However, the privacy protection required by the feature split makes the solution more complicated. We observe that, unlike cross-silo horizontal settings, the system performance difference is significant for frameworks implementing various vertical algorithms. Particularly, when considering the characteristics of computation, communication cost, and resource consumption, different algorithms provide clear trade-offs.

As discussed in Table \ref{tab:framework}, the supported model types in the vertical setting include regression, NN, and tree-based models. Here we compare the system performance to train the same model with different algorithms. The first step is to find candidates that provide the required training functionalities.
Fedlearner, FedTree, and FATE implement the distributed training of tree-based models, while Crypten, PaddleFL, FATE, and Fedlearner support the training of regression and NN-based models. 
However, in practice, for PaddleFL, we fail to run the sMPC example following the official instructions in its latest release 1.2.0 and its split learning feature is removed. 
Similarly, the official distributed deployment tool for FATE failed to be installed on standard AWS servers.
While Fedlearner supports distributed training for tree-based models, it only provides the network with one hidden layer for split learning off-the-shelf thus is not general to run any MLP for a fair comparison.

\begin{table}[h]
\centering
\resizebox{\columnwidth}{!}{
\begin{tabular}{l|cc|c}
\toprule
& \textbf{\makecell{Fedlearner\\SecureBoost\\GBDT}} & \textbf{\makecell{FedTree\\SecureBoost\\GBDT}} & \textbf{\makecell{CrypTen\\sMPC\\3-layer-MLP}}\\
\hline
Time & 9513.9s & 456.0s & 688.1s \\
Total comm & 1.33G & 1.46G & 66.15G \\
Memory & 0.69G, 0.40G & 2.16G, 0.64G & 0.38G, 0.38G \\
Final perf & 81.41\% & 81.89\% & 78.86\% (10 epochs) \\
\hline
\end{tabular}%
}
\caption{\small Performance comparison in vertical setup ``default\_credit\_vertical''. Memory consumption for both participants are reported.}
\label{tab:sysperfvertical}
\end{table}

Table \ref{tab:sysperfvertical} provides a performance comparison. We record the total transmitted bytes from all participants and sum it up to measure the communication cost. As shown in the table, there is a huge system performance difference between different algorithm implementations. As indicated by the analysis of the log, we observe the pattern that Fedlearner is only consuming one CPU core on the machine, thus is slower than FedTree (which uses 4 to 8 cores) implementation but consumes less memory because of the smaller parallel workload. Between FedTree and CrypTen, we observe that sMPC-based NN training also only use one core, is generally slower, introduces significant communication cost, but consumes less memory. 
We conclude that existing tree-based vertical FL frameworks are more efficient than NN-based FL frameworks.

%% file: content/550_device_h.tex
\subsection{Cross-device horizontal setting: parallelism and sampling cause performance difference }

Cross-device setups often involve more participants and thus harder to deploy. While certain frameworks use local simulators acting as different clients to reduce the simulation cost, with \sysname, we run evaluations in the actual deployment environment with hundreds of machines. Here we present the training results on 178 machines (with potentially a few more aggregators) in Table \ref{tab:femnist}.

\begin{table}[h]
\centering
\resizebox{\columnwidth}{!}{%
\begin{tabular}{c|cc|c}
\toprule
& \textbf{\makecell{Flower}} & \textbf{\makecell{FedML}} & \textbf{\makecell{Fedlearner}}\\
\hline
Time & 57.4s & 48.4s & 66.2s \\
Total comm & 0.62G & 0.58G & 0.83G \\
Memory & 2.12G, 1.64G & 1.18G, 1.08G & 1.82G, 1.63G \\
\hline
\end{tabular}%
}
\caption{\small System performance comparison for training a LeNet in setup ``femnist''. Memory consumption on aggregator (left) and client (right) nodes in each cell are reported separately. %
}
\label{tab:femnist}
\end{table}

\begin{table}[!hbtp]
\centering
\begin{tabular}{cc|c}
\toprule
\textbf{\makecell{Flower}} & \textbf{\makecell{FedML}} & \textbf{\makecell{Fedlearner}}\\
\hline
82.13\% & 80.40\% & 69.06\% \\
\hline
\end{tabular}%
\caption{\small Model performance comparison for training LeNet in setup “femnist”.}
\label{tab:femnistmodel}
\end{table}

For Flower and FedML, we run 25 epochs sampling 20\% clients each time to train a LeNet model. However, Fedlearner does not support the sub-sampling of the clients. To align the training procedure for a preliminary comparison, we run 5 epochs of training on Fedlearner targeting all the clients. Table \ref{tab:femnist} summarizes the results. We note that FedML achieves better system performance than Flower, probably due to different sampling procedures in its implementation. In addition, although Fedlearner handles the same total workload with similar time consumption, it should have completed faster due to higher parallelism in its 5 epochs. 
Considering that a complete training would require at least 2000 epochs, the absolute system performance difference could be significant in real-world deployment.

In terms of model performance, we present the result in Table \ref{tab:femnistmodel} after more training rounds: 2000 rounds for Flower and FedML, and 400 rounds for Fedleaner. Fedlearner's performance is significantly lower due to low training efficiency caused by gradient descent on data from all clients.

To better understand the reason behind the model performance difference between Flower and FedML, we investigated their implementations and concluded that the difference is mainly due to the different sampling procedures in the two frameworks.
In Flower, each client uses the same fixed set of instances for training across different epochs, while the server samples different subsets of clients in each epoch. On the other hand, in FedML, although all clients are connected to the server, only a fixed subset of clients is sampled for all epochs, and each client samples different sets of instances from the dataset for simulation.
Although both cases calculate model gradients from a random subset of instances sampled from the dataset, the mathematical procedure for selection is not the same.

To conclude, in cross-device settings, for framework selection one should consider the trade-off between the sampling method (which has potential influences on model performance) and system performance.

%% file: content/600_discuss.tex
\section{Discussion and future work}
\label{sec:discuss_future_work}

With \sysname, framework developers can easily comprehend potential bottlenecks in development.  Moreover, based on our observations (Table \ref{tab:framework}) and on experiments (Section \ref{sec:exp}), \sysname provides framework recommendations for the given FL setup. Specifically, FL practitioners can first analyze the qualitative requirement of the use case and narrow down the scope with Table \ref{tab:framework}. Then referring to example setups with similar settings discussed in Section \ref{sec:exp} to gain performance insights, they can start trying out different options conveniently with \sysname. Furthermore, the procedure could be automated with additional modeling. As an example shown in Figure \ref{fig:intuition}, we built a default decision tree\footnote{\url{https://bit.ly/unifed-tree}} in \sysname that uses interactive questions to help select frameworks based on our findings.

There are the following future directions to further improve \sysname: (1) we expect more datasets can be incorporated into \sysname as the FL studies grow, especially for vertical FL. (2) We will periodically check the latest and representative FL frameworks (e.g.,  {OpenFL}~\cite{openfl}, 
 {FLSim}~\cite{flsim},  {OpenFed}~\cite{chen2021openfed}
 ) and include them into \sysname. 
(3) We may evaluate the fairness, incentives  {and personalization} of FL frameworks when there are enough frameworks  {supporting these lines of research.}
(4) We plan to launch open competitions to  {make existing frameworks more secure and efficient.}

%% file: content/appendix/all.tex
\input{content/appendix/rebuttal}  %
\input{content/appendix/docreq}
\input{content/appendix/codepatch}

\input{content/appendix/moreexp}

%% file: content/appendix/rebuttal.tex
\section{Framework versions}
\label{sec:exp-details}

\paragraph{Versions} We specify the versions of the evaluated frameworks in Table~\ref{tab:version}. %
We use the corresponding ML backend for each framework as listed in Table~\ref{tab:framework}. 
For the frameworks supporting more than one ML backends, we choose TensorFlow for FATE because almost all of its official examples are using TensorFlow; we use PyTorch in Flower because covers the most examples provided by Flower; for FederatedScope and FedScale, we use PyTorch because it is their default supports, and their tutorials also use PyTorch. 

\begin{table*}[!htbp]
    \centering
\resizebox{0.55\textwidth}{!}{%
\begin{tabular}{lccccccc}
\toprule
 \textbf{\makecell{Framework}} & \textbf{\makecell{Git commit}}    \\
\hline
\multicolumn{1}{c}{FATE}  & \href{https://github.com/FederatedAI/FATE/tree/018d051f06298cd01aec957d569ff5760ff0070e}{018d051f06298cd01aec957d569ff5760ff0070e}  &  \\
\multicolumn{1}{c}{FedML} &  \href{https://github.com/FedML-AI/FedML/tree/a8b59a6346bf548dc66bd2266af5070ee15db4eb}{a8b59a6346bf548dc66bd2266af5070ee15db4eb}  \\
 \multicolumn{1}{c}{PaddleFL}& \href{https://github.com/PaddlePaddle/PaddleFL/tree/e949f194aec03d1d2e26530f6bf6f4e83026eb2d}{e949f194aec03d1d2e26530f6bf6f4e83026eb2d} \\
\multicolumn{1}{c}{Fedlearner} & \href{https://github.com/bytedance/fedlearner/tree/cd33b12614b77ae45331ce0e88077ca773b98d42}{cd33b12614b77ae45331ce0e88077ca773b98d42} \\
\multicolumn{1}{c}{FederatedScope} & \href{https://github.com/alibaba/FederatedScope/tree/5b2bb38675f3ab91b6282e0efb158a28ce6aa16d}{5b2bb38675f3ab91b6282e0efb158a28ce6aa16d}  \\
\multicolumn{1}{c}{TFF} & \href{https://github.com/tensorflow/federated/tree/bf81fc5d4386ac966b3b4e3d75403910507ea51a}{bf81fc5d4386ac966b3b4e3d75403910507ea51a}  \\
\multicolumn{1}{c}{Flower}  & \href{https://github.com/adap/flower/tree/d1eb90f74714a9c10ddbeefb767b56be7b61303d}{d1eb90f74714a9c10ddbeefb767b56be7b61303d}  &  \\ 
\multicolumn{1}{c}{FLUTE}   & \href{https://github.com/microsoft/msrflute/tree/9df34a8bfdbdbe768f4a3bbf6fe4c7284fede43c}{9df34a8bfdbdbe768f4a3bbf6fe4c7284fede43c}  &  \\ 
\multicolumn{1}{c}{FedScale} &  \href{https://github.com/SymbioticLab/FedScale/tree/d99f0cc2beb726c58294cc51af75fe9d8b163242}{d99f0cc2beb726c58294cc51af75fe9d8b163242}  &  \\ 
\multicolumn{1}{c}{CrypTen} &  \href{https://github.com/facebookresearch/CrypTen/tree/891fa4709e4849ff50ace6933f20375b04b3f722}{891fa4709e4849ff50ace6933f20375b04b3f722}  &  \\ 
\multicolumn{1}{c}{FedTree} &  \href{https://github.com/Xtra-Computing/FedTree/tree/d945e7ddd6180d919ef1f57f36acbb0c0a1c5295}{d9c0b0896343938e75563d1a8f8f632bd12ecc82}  &  \\ 
\hline
\end{tabular}%
}
\caption{\small {Version of the evaluated frameworks.}}
\label{tab:version}
\end{table*}

%% file: content/appendix/docreq.tex
\section{Survey on FL frameworks}
\label{sec:docreq}

\paragraph{Survey design}
We designed a questionnaire form with 18 
questions reflecting the framework support from the perspectives of model types, deployment, privacy protection, and utility. The original form is available at \url{https://bit.ly/unifed-survey}.

Specifically, for both horizontal FL and vertical FL, we consider the model supports for regression, neural networks, and tree-based models. 
Regarding the support for deployment, we consider various scenarios such as single-host simulation, multi-host simulation on a single machine, and multi-host simulation on different machines, etc. 
In terms of network protocols, our considerations include gRPC, MPI, and Torch Distributed.
When it comes to privacy protection, we assess the following aspects:
\begin{itemize}
    \item  The extent of client access to model parameters (e.g., completely hidden from the clients, partial disclosure, full access).
    \item  Whether clients have access to model gradients/updates from other clients.
    \item  Whether the aggregator has access to individual model gradients/updates.
    \item  Whether the framework supports private training mechanisms (e.g., DP-SGD, perturbed gradient, perturbed model update).
    \item  Whether the framework provides formal differential privacy (DP) guarantees for private training, such as user-level DP and sample-level DP.
\end{itemize}

Regarding utility, we take into account several factors. These include whether the framework provides its own (probably naive) test and performance benchmarks, whether it supports GPU, and its compatibility with various types of neural networks. We also consider the framework's support for different machine learning optimizers and its compatibility with popular ML backends such as PyTorch, TensorFlow, and JAX.
Additionally, we consider the usability features of different frameworks, particularly their documentation support. This involves evaluating the availability and quality of documentation provided for the frameworks. Concretely, 
\begin{enumerate}  [noitemsep,nolistsep]
    \item A \textit{tutorial} should give a step-by-step explanation for code writing, leading the user from the installation to a first successful end-to-end FL training procedure. 
    \item \textit{Example source code} files should demonstrate the use of framework APIs and can serve as boilerplate for supporting new but similar application scenarios.
    \item Fine-grained \textit{API documentation} should exhaustively explain the available configuration options to handle different FL scenarios in different network topologies and train different types of models. 
\end{enumerate}

We sent out the questionnaires to the maintainer teams of all 11 frameworks and received 5 official replies covering around a hundred critical answers.
Based on the survey and verified by our integration experience, we compare the functionality support of different frameworks in \Cref{tab:framework} and the usability features of different frameworks in \Cref{tab:usability}.

\input{content/appendix/support_alg}

%% file: content/appendix/support_alg.tex
\paragraph{Supported algorithms}

We compare the algorithm implementations, possible client failures, and asynchronous aggregation supports of FL frameworks in Table \ref{tab:algorithm}. For the horizontal FL setting, FedAvg is the only algorithm that is supported by all frameworks except for specialized frameworks. That is why we only evaluate the horizontal settings on FedAvg for RQ1. To handle data heterogeneity, some frameworks (FedML, FederatedScope, TFF, FedScale) support FedProx~\cite{li2020federated} or FedNova~\cite{wang2020tackling}
. For adaptive server optimizations, FedML, FederatedScope, Flower and FedScale support FedOpt~\cite{reddi2020adaptive}, FedAdam~\cite{reddi2020adaptive}, FedAdagrad~\cite{reddi2020adaptive} or FedYogi~\cite{reddi2020adaptive}, which enable use of adaptive learning rates without increase in client storage or communication costs, and ensure compatibility with cross-device FL. In contrast to the wide range of supports on horizontal settings, vertical FL is only supported by a part of the frameworks, and there is no vertical algorithm that is universally supported by all these frameworks. For example, Fedlearner only supports MLP with one hidden layer for split learning, while FederatedScope only supports regression.

\begin{table*}[!htbp]
    \centering
\renewcommand{\arraystretch}{1.5}
\resizebox{\textwidth}{!}{%
\begin{tabular}{lccccccc}
\toprule
\textbf{\makecell{Type}} & \textbf{\makecell{Framework}} & \textbf{\makecell{Horizontal Algorithms}} & \textbf{\makecell{Vertical Algorithms}}   &  \multirow{2}{*}{\textbf{\makecell{Possible client\\failures support}}} & \multirow{2}{*}{\textbf{\makecell{Asynchronous\\aggregation}}}     \\\\
\hline
\multicolumn{1}{c}{\multirow{10}{*}{\makecell{All-in-one\\frameworks}}} & \multicolumn{1}{c}{FATE} & {\makecell{Secure Aggregation~\cite{bonawitz17practical}, Paillier Encryption~\cite{paillier1999}, \\ HistSecAgg\cite{bonawitz2016practical} }}& {\makecell{ VFL-LR~\cite{hardy17private}~\cite{chen20when}, VFL-NN~\cite{zhang18gelu}~\cite{zhang20additively}, \\ SecureBoost~\cite{cheng2021secureboost} }}& \yes & \yes \\
\cline{2-6}
\multicolumn{1}{c}{} & \multicolumn{1}{c}{FedML} & {\makecell{FedAvg~\cite{mcmahan2017communication}, FedOpt~\cite{reddi2020adaptive}, FedNova~\cite{wang2020tackling}, \\ Decentralized FL,\\Hierarchical FL (Distributed computing,\\Mobile/IoT computing, etc)}} & VFL-LR~\cite{hardy17private} & \no & \no  \\
\cline{2-6}
\multicolumn{1}{c}{} & \multicolumn{1}{c}{PaddleFL} & FedAvg~\cite{mcmahan2017communication} & Two-party PrivC~\cite{he2019privc}, Three-party ABY3~\cite{mohassel2018aby3} & \no & \yes \\
 \cline{2-6}
\multicolumn{1}{c}{} & \multicolumn{1}{c}{Fedlearner} & FedAvg~\cite{mcmahan2017communication} & Two-party split learning~\cite{vepakomma2018split} & \yes & \no \\
\cline{2-6}
\multicolumn{1}{c}{} & \multicolumn{1}{c}{FederatedScope} & {\makecell{FedAvg~\cite{mcmahan2017communication}, Ditto~\cite{li2021ditto}, FedEM~\cite{marfoq2021federated}, \\ FedOpt~\cite{reddi2020adaptive}, FedProx~\cite{li2020federated},\\FedSage+~\cite{zhang2021subgraph}, NbAFL~\cite{wei2020federated}}} & Secure federated logistic regression~\cite{hardy2017private}
& \yes & \yes  \\\cline{2-6}
& \multicolumn{1}{c}{TFF} & {\makecell{FedAvg\cite{mcmahan2017communication}, FedProx\cite{li2020federated}, \\ FedSGD\cite{mcmahan2017communication}, Kmeans-clustering }} & \yes* & \yes & \no \\
\hline
\multicolumn{1}{c}{\multirow{4}{*}{\makecell{Horizontal-only\\frameworks}}} 
& \multicolumn{1}{c}{Flower} & {\makecell{FedAvg\cite{mcmahan2017communication}, FedAdam\cite{reddi2020adaptive}, FedAdagrad\cite{reddi2020adaptive},\\FedOpt\cite{reddi2020adaptive}, FedYogi\cite{reddi2020adaptive}}} & /& \yes & \no  \\
 \cline{2-6}
\multicolumn{1}{c}{} & \multicolumn{1}{c}{FLUTE} & FedAvg\cite{mcmahan2017communication}, DGA\cite{dimitriadis2021dynamic} & / & \yes & \no \\
 \cline{2-6}
\multicolumn{1}{c}{} & \multicolumn{1}{c}{FedScale} & {\makecell{FedAvg\cite{mcmahan2017communication}, FedProx\cite{li2020federated}, \\ FedYogi\cite{reddi2020adaptive}, q-FedAvg\cite{q-fedavg} }} & / & \yes & \yes \\
\hline
\multicolumn{1}{c}{\multirow{2}{*}{\makecell{Specialized\\frameworks}}} & \multicolumn{1}{c}{CrypTen} & / & sMPC\cite{crypten2020} & \no & \yes  \\
\cline{2-6}
\multicolumn{1}{c}{} & \multicolumn{1}{c}{FedTree} & HistSecAgg\cite{bonawitz2016practical} & SecureBoost\cite{cheng2021secureboost} & \no & \no \\
\hline
\end{tabular}%
}
\caption{{Algorithm implementations for each framework.}}
\label{tab:algorithm}
\end{table*}

%% file: content/appendix/codepatch.tex
\section{Code patch details in each FL framework}
\label{sec:patch}

FATE
provides a service interface named FATE-Flow which allows FL training task submission in the format of pipelines. We write Python scripts using FATE-Pipeline to construct various FL training pipelines. To run evaluation scenarios from our scenario loader, we upload the corresponding data records with pipeline-upload to FATE-Flow. For comprehensive logging for global analysis, we add tracking in the pipeline Python script and also inject code patches to FATE libraries.

FedML
provides the flexibility to implement custom data-loading modules and model architectures. It also uses a configuration file to indicate the dataset, the model architecture, and the training parameters. Therefore, we are able to adapt our configuration file to FedML's format and use its Python script to start the training. We track all metrics with our code patch.

PaddleFL
provides an interface for defining models. We modify its demo Python scripts to adapt our scenarios and calculate the target model performance metrics. To get the system performance metrics, we inject our logging code to its core. 
We also shorter the sleep time after each epoch and wait action in this framework from 3s/5s to 0.2s to speed up our evaluation procedure. We find the split learning algorithm for vertical scenarios in this framework does not work because the code generation function for the grpc proto file is been commented in its setup.py, and the demo code can not run normally. And we also can not run the demo code for mpc in its latest official docker image\footnote{\tiny Digest:sha256:06ea5f85ab5d048740bc8feffc74058cc32cccba422f346a1621f10300dba410}.

Fedlearner
provides an interface to define both the model and metrics in its FedAvg algorithm implementation. We set our models and expected metrics in our Python script to test our scenarios. In its tree-based model algorithm, it provides parameters to set the necessary training parameters. However, its split learning model is hard to customize. By default, it only provides one-layer networks and it needs significant efforts to support three layers networks. Similar to FedML, We track all metrics we need by our code patch in it. %

TFF
provides a wrapped interface for simulation on a single host and we can get the model performance metrics and system performance metrics we need from it directly. So we write the corresponding Python scripts.

Flower
requires self-defined training and evaluating procedures. Thus, we write our evaluation code to build different evaluation scenarios and get the model performance metrics through it. And we also add a code patch to get the system performance metrics from it.

FLUTE
starts training procedures from configuration files. So we write a script to convert our evaluation scenario and dataset into its format and define our model in the configuration file. In addition, we inject our logging system into its codebase to get both the model performance metrics and system performance metrics.

CrypTen
provides a PyTorch-like Python API, so we write a Python script to define our model and pass our data. And we also use our Python script to start the evaluation and get the model performance metrics and parts of system performance metrics like computation cost. To get the communication cost, we patch our logging system into its statistics-related module.

FedTree
is a framework written in C++ and also supports starting from configuration files. We write a script to convert the evaluation scenario to its configuration file and start the evaluation. Because of its C++ backend, it's hard to directly inject our logging patch into it, so we use a script to convert its log information into our logging files.

{FedScale} 
{is a scalable and extensible open-source federated learning (FL) engine and benchmark. We mainly work with its code of model and dataset while injecting our logging system into it.}

{FederatedScope}
{provides some examples of dataset and model customization, where we write specifically-formatted code files in a specific directory. These datasets and models are registered into the framework when the configuration calls them. We also write a script to convert our scenarios and parameters into its configuration format, which is required to start training. For unified logging, we leverage a code patch to inject our logging system into the original code of the framework.}

%% file: content/appendix/moreexp.tex
\section{Experimental output and additional results}
\label{sec:addexp}

In the main paper, we only cover representative results to demonstrate our findings. In this section, we demonstrate the evaluation metrics supported by \sysname and present the additional experimental results that are not covered in the main paper.

\subsection{Metrics}

Our logging output covers various metrics including model performance and system performance.
To get the performance metrics such as the time consumption of computations and communications, the size of communication data, and the model performance, we need to track the execution procedure of the target framework. 
We inject our logging system into the target FL framework codebase to track the timestamp and meta-information about the beginning and the ending of certain events.
Here in Figure~\ref{fig:log} is an example of the log format.

\begin{figure*}
\tiny
\begin{verbatim}
{"flbenchmark": "start", "timestamp": 1653923858.0422723, "agent_type": "aggregator"}
{"event": "training", "action": "start", "timestamp": 1653923858.04346, "metrics": {}}
{"event": "training.0", "action": "start", "timestamp": 1653923858.0435393, "metrics": {}}
{"event": "computation.0", "action": "start", "timestamp": 1653923860.8540547, "metrics": {}}
{"event": "computation.0", "action": "end", "timestamp": 1653923861.450064, "metrics": {"flop": 123, "loss": 0.8}}
{"event": "communication.1.0", "action": "start", "timestamp": 1653923861.4501162, "metrics": {}}
{"event": "communication.1.0", "action": "end", "timestamp": 1653923861.450492, "metrics": {"byte": 1234}}
{"event": "communication.2.1", "action": "start", "timestamp": 1653923861.4505105, "metrics": {}}
{"event": "communication.2.1", "action": "end", "timestamp": 1653923861.450583, "metrics": {"byte": 1234}}
{"event": "training.0", "action": "end", "timestamp": 1653923861.4505887, "metrics": {}}
{"event": "training.1", "action": "start", "timestamp": 1653923861.4506032, "metrics": {}}
...
{"event": "training.3", "action": "end", "timestamp": 1653923867.9962113, "metrics": {}}
{"event": "training", "action": "end", "timestamp": 1653923867.996218, "metrics": {}}
{"event": "model_evaluation", "action": "start", "timestamp": 1653923867.9962301, "metrics": {}}
{"event": "model_evaluation", "action": "end", "timestamp": 1653923868.0964506, "metrics": {"accuracy": 99.9}}
{"flbenchmark": "end", "timestamp": 1653923868.0964868}
\end{verbatim}
\normalsize
\caption{Logging format.}
\label{fig:log}
\end{figure*}

\subsection{Model performance}
While \sysname supports 15 different FL scenarios, in addition to ``give\_credit\_horizontal'' presented in Section~\ref{sec:csh}, we select four scenarios to compare the model performance across different frameworks covering finance (``default\_credit\_vertical''), healthcare (``breast\_horizontal'' and ``breast\_vertical''), and image classification (``femnist''). The results are presented in Table~\ref{tab:breast_horizontal_mp} to Table~\ref{tab:tree_mp} (we omit the frameworks if they do not support the corresponding settings). From the results, we observe that the frameworks achieve similar model performance given the same model in most cases since they have the same underlying algorithms. Also, we have the following interesting findings.
\begin{itemize}
    \item FATE has limited support in the cross-device setting (e.g., femnist). This limitation arises from a file naming issue within FATE, which prevents it from scaling to handle hundreds of clients effectively. Also, FATE has a relatively low performance when using logistic regression in ``breast\_horizontal'', which might be relevant to its default early-stop behavior triggered by convergence.
    \item PaddleFL consistently exhibits lower model performance compared to other frameworks in the femnist scenario. This discrepancy can be attributed to PaddleFL's utilization of a distinct machine learning backend, PaddlePaddle~\cite{ma2019paddlepaddle}.
    \item Vertical FL, unlike horizontal FL, encounters limited support within existing FL frameworks. Furthermore, unlike the widely used FedAvg algorithm in horizontal FL and lossless federated training of GBDT, there is no commonly agreed de-facto approach for vertical federated training of neural networks. These factors contribute to performance differences between frameworks.
\end{itemize}

In summary, \sysname offers framework developers a convenient tool for comparing model performance and identifying anomalies, facilitating debugging and framework improvements. Additionally, to minimize the effort of trying multiple frameworks, users can refer to our recommended framework selection tree for choosing the most suitable framework for a given scenario.

\begin{table*}
\centering
\resizebox{\textwidth}{!}{%
\begin{tabular}{c|ccccccccccc}
\toprule
\textbf{\makecell{Model}}      & \textbf{\makecell{FATE}}              & \textbf{\makecell{FedML}}             & \textbf{\makecell{PaddleFL}}     & \textbf{\makecell{Fedlearner}}     & \textbf{\makecell{FederatedScope}}    & \textbf{\makecell{TFF}}               & \textbf{\makecell{Flower}}             & \textbf{\makecell{FLUTE}}             \\ \hline
\multicolumn{1}{l}{Regression} & 0.624$\pm$0.041 & 0.982$\pm$0.002 & 0.981$\pm$0.001 &0.986$\pm$0.001 & 0.981$\pm$0.003 & 0.988$\pm$0.001 & 0.982$\pm$0.002  & 0.982$\pm$0.005 \\ 
 \multicolumn{1}{l}{SLP}      & 0.984$\pm$0.002 & 0.987$\pm$0.003 & 0.982$\pm$0.002 &0.988$\pm$0.001 & 0.986$\pm$0.003 & 0.983$\pm$0.001 & 0.987$\pm$0.001 & 0.985$\pm$0.004 \\ 
\multicolumn{1}{l}{3-layer MLP}      & 0.986$\pm$0.002 & 0.987$\pm$0.001 & 0.982$\pm$0.002 &0.986$\pm$0.001 & 0.987$\pm$0.001 & 0.986$\pm$0.001 & 0.987$\pm$0.003  & 0.986$\pm$0.003 \\ \hline
\end{tabular}%
}
\caption{\small Model performance comparison across different frameworks in the setup ``breast\_horizontal''.}
\label{tab:breast_horizontal_mp}
\end{table*}

\begin{table*}
\centering
\resizebox{\textwidth}{!}{%
\begin{tabular}{c|cccccccccc}
\toprule
\textbf{\makecell{Model}}    & \textbf{\makecell{FedML}}             & \textbf{\makecell{PaddleFL}}          & \textbf{\makecell{FederatedScope}}    & \textbf{\makecell{TFF}}               & \textbf{\makecell{Flower}}             & \textbf{\makecell{FLUTE}}    & \textbf{\makecell{FedScale}}           \\ \hline
\multicolumn{1}{l}{Regression} & 0.044$\pm$0.011 & 0.066$\pm$0.008 & 0.004$\pm$0.003 & 0.064$\pm$0.014 & 0.038$\pm$0.004 & 0.037$\pm$0.002 & 0.127$\pm$0.017 \\
\multicolumn{1}{l}{SLP}      & 0.647$\pm$0.004 & 0.533$\pm$0.035 & 0.655$\pm$0.006 & 0.676$\pm$0.007 & 0.652$\pm$0.011 & 0.602$\pm$0.008 & 0.674$\pm$0    \\
\multicolumn{1}{l}{3-layer MLP}      & 0.699$\pm$0.005 & 0.594$\pm$0.015 & 0.698$\pm$0.006 & 0.757$\pm$0.062 & 0.690$\pm$0.016 & 0.589$\pm$0.009 & 0.637$\pm$0.027 \\ 
\multicolumn{1}{l}{LeNet}      & 0.808$\pm$0.006 & 0.770$\pm$0.011 & 0.834$\pm$0.001 & 0.816$\pm$0.059 & 0.816$\pm$0.005 & 0.744$\pm$0.027 & 0.633$\pm$0.015 \\
\hline
\end{tabular}%
}
\caption{\small Model performance comparison across different frameworks in the setup ``femnist'' sampling 20 clients.}
\label{tab:femnist_mp}
\end{table*}

\begin{table*}
\centering
\begin{tabular}{c|cccc}
\toprule
\textbf{\makecell{Model}}      & \textbf{\makecell{FATE}}              & \textbf{\makecell{FedML}}             & \textbf{\makecell{FederatedScope}}    & \textbf{\makecell{CrypTen}}           \\ \hline
\multicolumn{1}{l}{Regression} & 0.967$\pm$0.011 & 0.988$\pm$0.002 & 0.937$\pm$0.034 & 0.996$\pm$0 \\
\multicolumn{1}{l}{SLP}      & \no     & 0.993$\pm$0.001 & \no     & 1.0$\pm$0 \\
\multicolumn{1}{l}{3-layer MLP}      & 1$\pm$0           & 0.995$\pm$0.001 & not supported     & 1.0$\pm$0     \\ \hline
\end{tabular}%
\caption{\small Model performance comparison across different frameworks in the setup ``breast\_vertical''.}
\label{tab:breast_vertical_mp}
\end{table*}

\begin{table*}
\centering
\begin{tabular}{c|cccc}
\toprule
\textbf{\makecell{Model}}      & \textbf{\makecell{FATE}}                             & \textbf{\makecell{FedML}}             & \textbf{\makecell{FederatedScope}}                 & \textbf{\makecell{CrypTen}}           \\ \hline
\multicolumn{1}{l}{Regression} & 0.696$\pm$0.003 & 0.605$\pm$0.012 & 0.689$\pm$0.010              & 0.705$\pm$0.001 \\
\multicolumn{1}{l}{SLP}     & \no                   & 0.663$\pm$0.003 & \multirow{2}{*}{\no} & 0.777$\pm$0.007 \\
\multicolumn{1}{l}{3-layer MLP}      & 0.741$\pm$0.005 (1 epoch)     & 0.970$\pm$0.011 (50 epochs) &                                & 0.788$\pm$0.001 \\\hline
\end{tabular}%
\caption{\small Model performance comparison across different frameworks in the setup ``default\_credit\_vertical''. We run FATE for only one epoch due to its slow training speed ($>$6 hours per epoch). We run FedML for 50 epochs since it achieves only 0.5 AUC given 10 epochs.}
\label{tab:default_credit_vertical}
\end{table*}

\begin{table*}
\centering
\begin{tabular}{c|c|ccc}
\toprule
\textbf{\makecell{Model}}   & \textbf{\makecell{Framework}} & \textbf{\makecell{breast\_horizontal}} & \textbf{\makecell{breast\_vertical}} & \textbf{\makecell{default\_credit\_vertical}} \\ \hline
\multirow{3}{*}{GBDT} & FATE      & 0.986$\pm$0  & 1$\pm$0          & 0.820$\pm$0             \\
                      & FedTree   & 1$\pm$0             & 1$\pm$0          & 0.814$\pm$0             \\ 
                      & Fedlearner   & \no             & 1$\pm$0          & 0.819$\pm$0             \\ \hline
\end{tabular}%
\caption{\small Model performance comparison between FATE and FedTree using GBDT across different setups. The training of GBDT is deterministic so the standard derivation is zero.}
\label{tab:tree_mp}
\end{table*}

\subsection{System performance}
\label{sec:grafana}

\begin{figure*}[!hbt]
\centering
    \includegraphics[width=\textwidth]{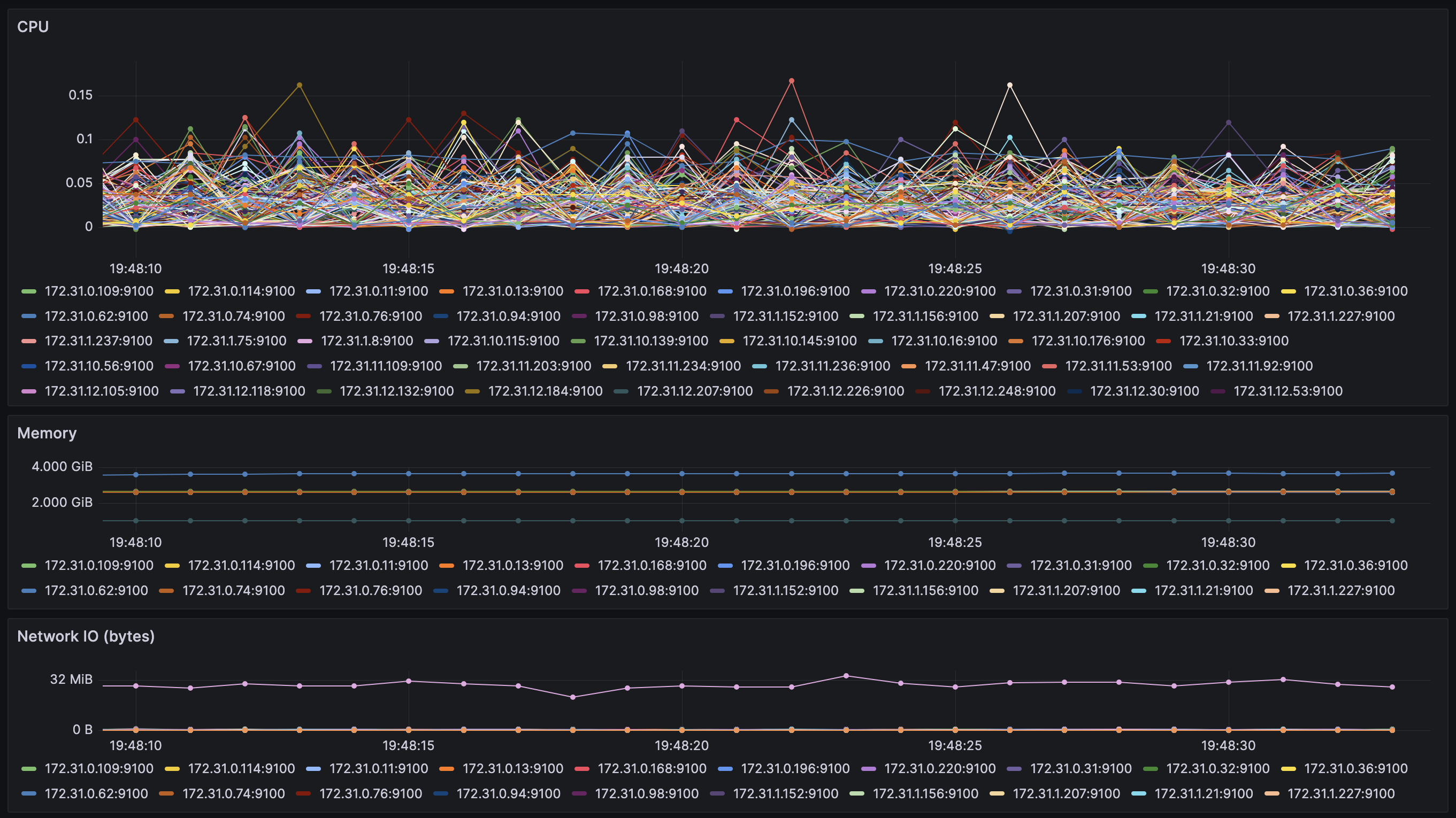}
      \caption{\small Example of CPU, Memory, and Network IO monitoring in FL training.}
    \label{fig:grafana}
\end{figure*}

\sysname enables live tracking for all evaluation nodes. In addition to the statistics reported in the main paper, \sysname also tracks the dynamics in the training procedure. In Figure \ref{fig:grafana}, we provide an example of using Grafana to illustrate the system performance tracking during the training procedure. We can directly observe the sampling procedure in Flower framework from CPU usage.

